# Gaussian Filter in CRF Based Semantic Segmentation

Yichi Gu[1,2], Qisheng Wu[1], Jing Li[1], Kai Cheng[1]

**Abstract**

Artificial intelligence is making great changes in academy and industry with the fast development of deep learning, which is a branch of machine learning and statistical learning. Fully convolutional network [1] is the standard model for semantic segmentation. Conditional random fields coded as CNN [2] or RNN [3] and connected with FCN has been successfully applied in object detection [4]. In this paper, we introduce a multi-resolution neural network for FCN and apply Gaussian filter to the extended CRF kernel neighborhood and the label image to reduce the oscillating effect of CRF neural network segmentation, thus achieve higher precision and faster training speed.

## 1. Introduction

Traditional machine learning algorithms and deep learning algorithms are two important methods for segmentation. Traditional algorithm extracts the object features by statistics, transformations, geometry etc., while deep learning algorithms automatically extract features by filters and deformations. A lot of deep learning neural networks are developed, e.g. deep Boltzmann network [11], deep belief network [12], deep convolutional network [16] etc. Several neural networks are combined, e.g. Auto Encoder [14], FasterRCNN [6], GAN [13] etc. The fusion of different kinds of neural networks better examine local and global, dominant and recessive properties with faster speed and higher accuracy.

In semantic segmentation, the network is optimized according to the label image with the segmentation information by assigning each pixel a class label. Fully convolutional neural network [1] was invested for semantic segmentation. It contains two parts: convolutional neural network (CNN) and the corresponding deconvolutional neural network (DCNN) to extract features and reconstruct image. Convolutional neural networks, such as VGG [15], Inception [6], Resnet [7], YoloII [5] win the first places in imaging competition. For three dimensional object detection, we adopt the multi-resolution structure and combinations of two dimensional filters in the convolutional neural network. Although CNN is a multi-resolution framework by sampling operation, branches with different length kernels are combined to strengthen the power of system.

CRF is a statistical model of maximum a posteriori with Gibbs distribution. Several methods are explored to solve CRF, e.g. independent conditional model, expectation maximization algorithm, mean field method, etc. By mean field approximation, the iterated solution can be represented by CNN, transferred to RNN and solved by network optimization. In this paper, we apply Gaussian filters to extend kernel function and the label image to reduce the noise caused by sharp edges with large gradient in the discrete data.

We will describe the FCN and CRF in session 2 followed by the adaption of Gaussian Filter in session 3. Experiments on CRF network for lung nodule segmentation are discussed in session 4 and conclusions will be drawn in session 5.

[1] TQ Medical Tech. Co.   [2] Contact Info.: yichig@163.com

## 2. Deep Learning Neural Network

2.1 Artificial Neural Network

ANN is a mathematical model of network approximation system inspired by the biological neural networks. It establishes and optimizes multi-layered or self-adjusted functions for detection, classification, segmentation etc.. A typical ANN is characterized by three points: (1) architecture specifies variables and topological relations of the network, such as component or transform weights, they level the strength of the passing signal; (2) activation function defines the message transform term to filter, re-scale and normalize the signal; (3) learning rule formulates the tuning operation of weights along the time. Neural networks have been used on a variety of tasks, including computer vision, speech recognition, social network filtering, playing games, medical diagnosis and in many other domains.

2.2 Convolution Neural Network

Convolutional neural network is a class of deep, feed-forward artificial neural network which has been successfully applied in the systems of image analysis and natural language processing.

CNN is characterized by the properties of local connection, weight sharing and down-sampling layers. Local connection and weight sharing reduce the quantity of parameter remarkably, so that the training complexity is decreased and the over-fitting problem is alleviated. Down-sampling also decreases the amount of parameters and predictions, and makes the network adapt to deformation which improves the generality of the network. CNN is an end to end filtration bank of convolutions, sampling, deformations, activations, normalizations etc.. The structure and component of CNN vary for different applications.

2.3 Inception and Resnet

Since the breakthrough of AlexNet, CNN is armed with deeper networks and wider layers which cause problems of computational complexity and gradient dispersion. In order to solve those problems, Google proposed a series of Inception-Resnet networks [7, 19, 20, 21] for image recognition. In the convolution layers, kernels of length (1,7) are applied; square convolution is factorized by directional convolutions; crosscut connections between layers are added. These improvements accelerate network calculation and increase the system's non-linearity. Inception-Resnet makes outstanding performance with large computation. Specified models are utilized in practice.

2.4 Fully Connected Neural Network

FCN was introduced in [1] for semantic segmentation. It connects CNN with deconvolutional network. The DCNN is composed of deconvolution operations and unpooling operations which precise the segmentation [8]. The convolution and deconvolution layers are fused together by adding to reuse the features while U-Net [9] concatenate the layers to segment the bio-medical image. The deconvolution network performs the super-resolution function and the whole network is concise and efficient. Their success in semantic segmentation shows the advantage of cross-layer fusion which is also applied in Resnet, Densely connected network etc..

2.5 Conditional Random Fields

For object detection, the plane or cubic convolution kernels may not be enough to extract the special object feature. CRF considers both pixel wise energy and the neighborhood energy which compensates for convolution. Following [2,3], we have the CRF as RNN framework.

Let *(I, L)* denote the pair of image and label and the conditional probability *P(L|I)* is modeled with Gibbs distribution of the form *P(L=x|I)= exp(-E(x|I))/Z(I)* where *E(x|I)* is the combination of unary energy and pairwise potentials as weighted Gaussians. Maximization of the conditional probability is equivalent to minimize energy *E(x|I)*. By mean-field approximation, *P(L=x|I)= Q(x)≈ ∏$_i$Q$_i$(x$_i$)*. The convolution network as the kernel of RNN is as follows.

(1) Initialization: $\quad Q_i(l) = \frac{1}{Z_i} exp(U_i(l)) \quad$ for all i

(2) Kernel operation: $\quad \tilde{Q}_i^{(m)}(l) = \sum_{j \neq i} k^{(m)}(f_i, f_j) Q_i(l)$

$$k(f_i, f_j) = w^{(1)} exp\left(-\frac{|p_i-p_j|^2}{2\theta_\alpha^2} - \frac{|I_i-I_j|^2}{2\theta_\beta^2}\right) + w^{(2)} exp\left(-\frac{|p_i-p_j|^2}{2\theta_\gamma^2}\right)$$

(3) Weighted filter operation: $\quad \check{Q}_i(l) = \sum_m w^{(m)} \tilde{Q}_i^{(m)}(l)$

(4) Compatibility transform: $\quad \hat{Q}_i(l) = \sum_{l' \in L} \mu(l, l') \check{Q}_i(l')$

(5) Add unary potential: $\quad \check{Q}_i(l) = U_i(l) - \hat{Q}_i(l)$

(6) Softmax normalization, while not determinate go to step (2).

In the above, $U_i(l)$ is the unary energy, *k(fi, fj)* is the kernel function where *fi* and *fj* are binary neighbors, *w$^{(1)}$,w$^{(2)}$* are the coefficients of appearance and smooth kernels, *μ(l,l')* are compatibility coefficients. After choosing appropriate parameters of kernels and initial values, the CNN optimizes the kernel coefficients and compatibility coefficients.

2.6 Combination of FCN and CRF

The semantic segmentation network FCN is connected with CRF by fully connection. The whole network actually outputs two predictions and the loss function is set to be the linear combination of the cross entropy of each prediction and the coefficient tunes the gradient scale on the parameter increment so that the system converges smoothly.

By considering the advantages of CNNs, we adopt the following model. Fig.1 shows the whole structure of the net which is composed of multi-blocks and dropout layer. Each multi-block is the concatenate combination of convolutions denoted in box and average pooling in diamond. The first four multi-blocks apply plane kernels while the last two multi-blocks use cubic kernels. Kernels of size 1-7 appear in all multi-blocks simultaneously. For different layers, the plane kernels are selected in different directions. For three dimensional images, plane kernel decomposition computes faster than cubic kernel and works better than the linear kernel decomposition.

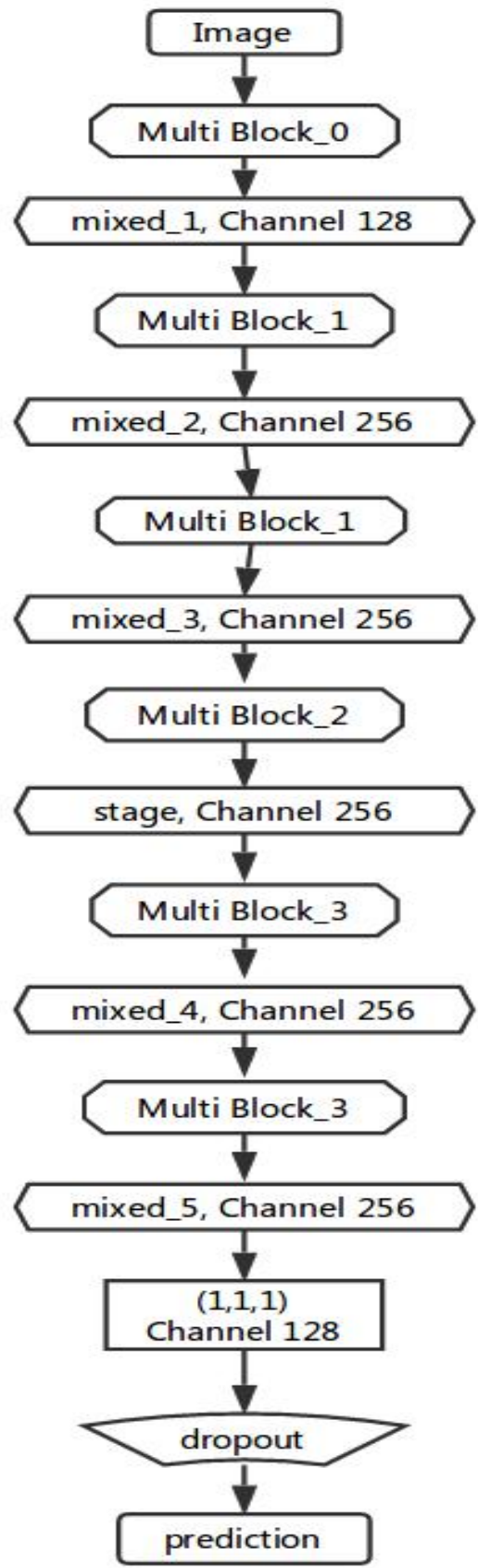

Fig. 1

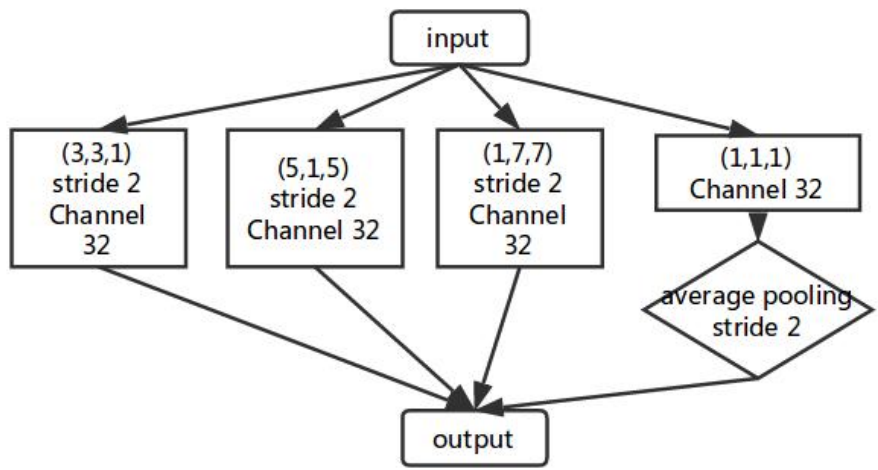

Fig. 2

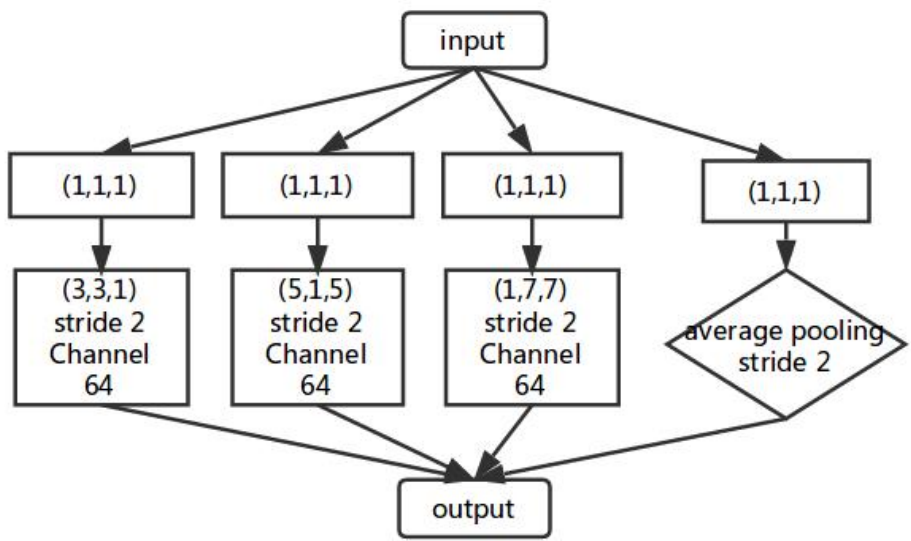

Fig. 3

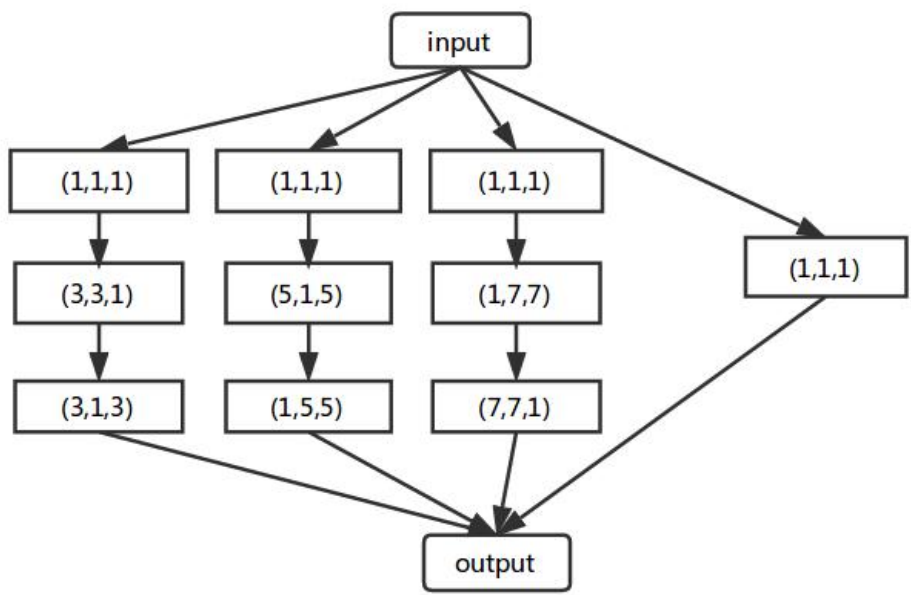

Fig. 4

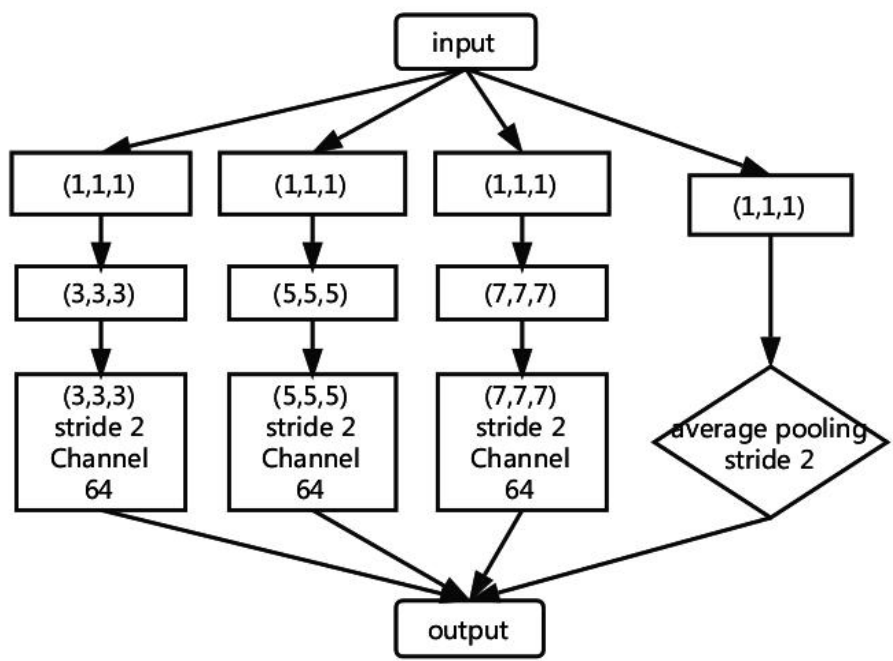

Fig. 5

## 3. Function of Gaussian Filter

3.1 Convolution and Filters

The convolution of two signals is the sliding sums of their multiplications. In frequency domain, its spectrum is the multiplication of the signals' spectra. The convolution in solid space is corresponded to the filtration in frequency space and the kernel signal acts as filter. Convolution has wide applications: linear time-invariant system representation, noisy system representation, edge extraction, smooth transform etc.. which computes linearly but adjusts the frequency spectrum. Convolution and deconvolution are a pair of important operations dealing with various problems in physics and engineering.

3.2 Gaussian Filter

Gaussian filter is expressed as the normal distribution which is the limiting distribution of the normalized sum of iid random variables. It is characterized by the center and variance. Due to the large quantity of similar individuals, a variety variables can be approximated by Gaussian variable or Gaussian process, such as noise, heat, quantum etc.. Gaussian filter has the highest value in the center and FWHM proportional to the standard deviation, decreasing rapidly in the area three times of standard deviation away from the center. With these properties, Gaussian filter is used as low pass filter for smoothing or denoising.

3.3 Applications of Gaussian Filter in Semantic Segmentation and CRF

In the network for semantic segmentation, loss function is the cross entropy of the resulting predictions and the label image. Label image is piece wise constant in different areas while the output prediction is a probabilistic image. The Gaussian smoothing on the label image reduces the absolute error and gradient on object boundary to enhance the precision.

In the CRF as CNN framework, Gaussian distribution is extended on the kernel, $g(|f_i-f_j|)k(f_i, f_j)$ where $g$ is the truncated weight function. The original six-neighborhood kernel function is extended and smoothed. Faster convergence is obtained since the scaled and fully connected neighborhood is applied. Variant extensions of CRF can be found in [17,18].

## 4. Experiments on Lung Nodule Detection

Our algorithm is implemented for lung nodule segmentation. The average speed is approximately 0.55s/step for Inception network and 1.2s/step for CRF using an Intel Xeon E5-1620 v3 CPU with 3.50GHz and a single NVIDIA GeForce GTX 1080Ti GPU. The proposed algorithm is tested on standard data set from Competition of Lung Nodule Analysis 2016 (LUNA16).

4.1 Gaussian Kernel Function

As described in [2], the pairwise edge potentials of our CRF models are defined by a linear combination of Gaussian kernels in feature space. We extend and re-scale the Gaussian kernel by evaluating the Gaussian kernel function on enlarged neighborhoods as below. Fig. 6 shows three modes of neighborhood: six neighborhoods, eighteen neighborhoods and twenty-six neighborhoods. We choose eighteen neighborhoods to construct binary energy function. Further more, scaled weights are distributed to different neighborhoods, such as one for six neighborhoods points and α for other twelve neighborhood points.

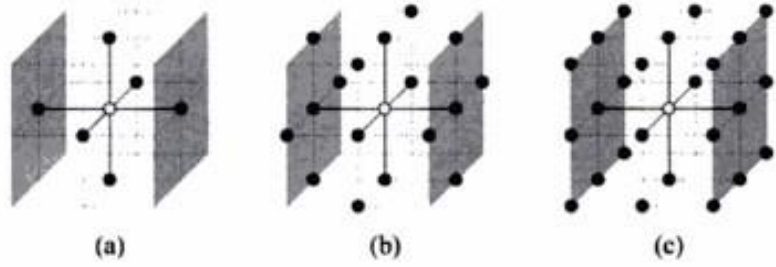

Fig. 6 different neighborhoods of 3D space

Three experiments are conducted under the same parameters: 10000 iterations of FCN network, 5000 iterations of FCN network plus additional 5000 iterations with six neighborhoods of CRF and 5000 iterations of FCN network plus additional 5000 iterations with eighteen neighborhoods of CRF.

|  | Loss | Pos. Prec. (%) | Neg. Prec. (%) |
|---|---|---|---|
| FCN(10000 steps) | 0.5530 | 72.8821 | 93.2632 |
| FCN(5000 steps) + 6-CRF(5000 steps) | 0.5428 | 72.5146 | 95.4866 |
| FCN(5000 steps) + 18-CRF(5000 steps) | 0.5250 | 77.5542 | 98.1018 |

Table 1

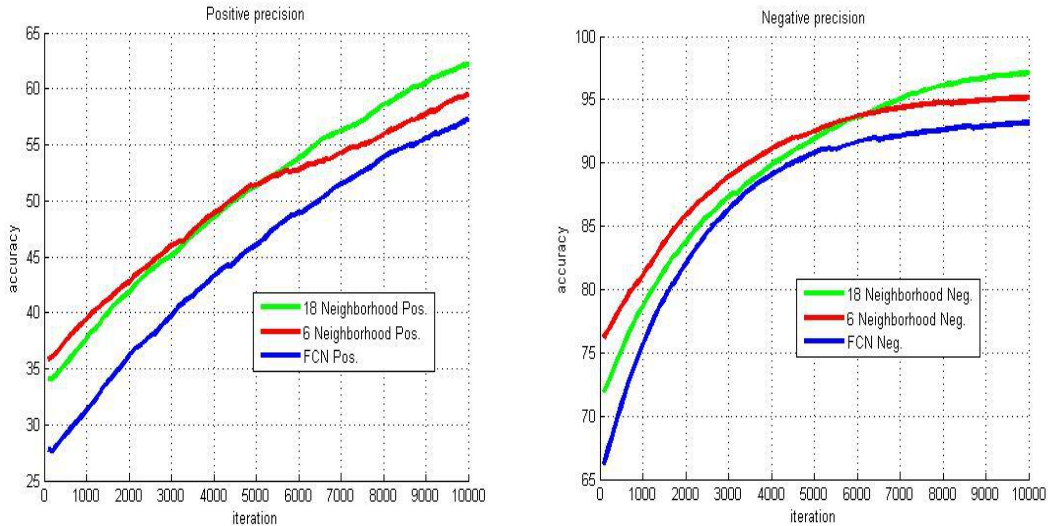

Fig. 7 Precision of FCN, six neighborhoods kernel and eighteen neighborhoods kernel

Table 1 compares the tests of three networks on the same data of LUNA16. Fig. 7 shows the precision of different neighborhoods. We obtain the following results: (1) with the same parameters and the same iterations, FCN network with CRF works better than FCN network according to the positive and negative precision; (2) CRF with eighteen neighborhoods kernel Gaussian function performs better than six neighborhoods kernel. This experiment shows the accuracy improvement of the CRF with extended Gaussian kernel in deep learning model.

4.2 Gaussian Filtered Labels

In semantic segmentation, the label image is a binary image, in which one represents nodule area and zero otherwise. Because of the sharp edges of nodule, ringing problem may be caused. Many methods can eliminate ringing effect, such as averaging, regularization, multi-filtration etc.. We apply the Gaussian filter to the label image and obtain a mask image which keeps value one on nodules, blurs on the edge and is zero otherwise.

Specifically, we apply the mask image on the computation of loss function by computing the cross entropy of prediction and the mask image, then multiplies the weighted label image. Table 2 compares the results of Gaussian filtered loss function and the original one. Fig. 8 shows the precision of different loss functions. According to the results of this table, Gaussian filter has little influence on the convolution-FCN network but shows changes on CRF network. Generally, the network can achieve better performance after filtering, which proves that Gaussian smoothing alleviates the oscillating effect caused by high frequency components in the label image.

The weighted label image is to be chosen in different situations. In lung nodule detection, it can be the weighted characteristic function of nodules. For further false positive reduction, it is set to be the weighted characteristic function of nodule and falsely classified nodule pixels.

|  | FCN(5000 iterations) | | | FCN + 18-CRF(5000 iterations) | | |
|---|---|---|---|---|---|---|
|  | Loss | Pos. Prec. (%) | Neg. Prec. (%) | Loss | Pos. Prec. (%) | Neg. Prec. (%) |
| Original objective function | 0.66 | 59.94 | 95.28 | 0.60 | 63.12 | 94.91 |
| Filtered objective function | 0.66 | 52.46 | 95.28 | 0.58 | 62.18 | 98.41 |

Table 2

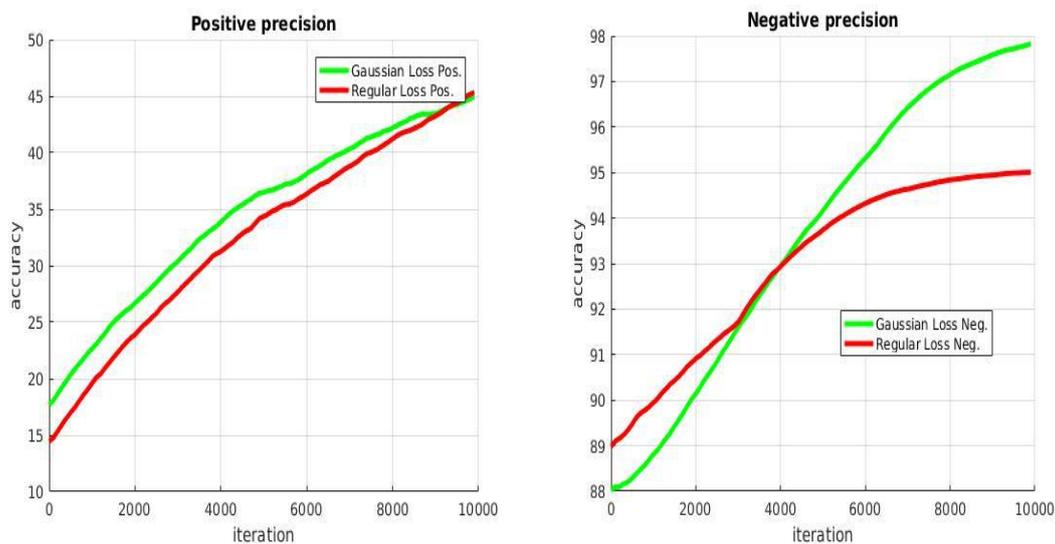

Fig. 8  Precision of Gaussian filtered Loss Function vs. Regular Loss Function

4.3 Training Method

For lung nodule detection project, we adopt the network described in session 2.6 in which a five-layer multi-CNN is used to construct FCN followed by CRF. In training stage, we use ADAM optimizer and 0.85 as keeping probability in dropout layer. As to network parameters, such as learning rate, the CRF weight and regularization parameters, they are all piece-wise constant along the iterations. As in FasterRCNN [10], FCN and CRF are iterated alternatively to improve the convergence of the system. Fig. 9 shows the contour of the detected nodules.

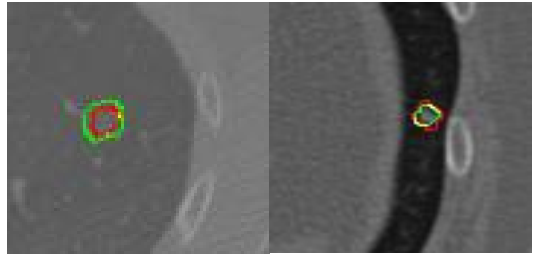

Fig. 9 Green: Detected Lung Nodule vs. Red: Ground Truth

## 5. Conclusions

From the above description and experiments, we conclude that the Gaussian filter on the extended CRF kernel function and the label image for semantic segmentation provides better precision. We also introduce a multi-box CNN model for imaging which decomposes cubic convolution to plane convolutions and combines kernels of different length in the same layer. These modifications work together with regularization, batch normalization, data augmentations to optimize the neural network system. Network parameters are also important to fine tune all these factors to achieve the best result.

## Reference

[1] Fully Convolutional Networks for Semantic Segmentation, Jonathan Long etc., CVPR2015, arXiv: 1411.4038

[2] Efficient Inference in Fully Connected CRFs with Gaussian Edge Potentials, Philipp Kr ahenb uhl etc., Oct. 2012, arXiv: 1210.5644

[3] Conditional Random Fields as Recurrent Neural Networks, Shuai Zheng etc., April 2016, arXiv: 1502.03240

[4] Efficient multi-scale 3D CNN with fully connected CRF for accurate brain lesion segmentation, Konstantinos Kamnitsas etc., Medical Image Analysis, 2017, 36(61-78)

[5] YOLO9000: Better, Faster, Stronger, Joseph Redmon etc. arXiv: 1612.08242v1, Dec. 2016

[6] Rethinking the Inception architecture for computer vision, Christian Sregedy etc. arXiv:1512.00567v3, Dec. 2015

[7] Inception-v4, Inception-ResNet and the Impact of Residual Connections on Learning, Christian Sregedy etc. Aug. 2016, arXiv: 1602.07261v2

[8] Learning Deconvolution Network for Semantic Segmentation, Hyeonwoo Noh etc., May 2015, arXiv: 1505.0436

[9] U-Net: Convolutional Networks for Biomedical Image Segmentation, Olaf Ronneberger etc., MICCAI 2015, arXiv: 1505.04597

[10] Faster R-CNN: Towards Real-Time Object Detection with Region Proposal Networks. Shaoqing Ren etc. arXiv: 1506.01497v3, Jan. 2016

[11] Reducing the Dimensionality of data with Neural Networks, G. E. Hinton, 28 July 2006, VOL 313 Science

[12] Introduction to Deep Learning and Case Study, Y. Li, 2017

[13] Unsupervised representation learning with Deep Convolutional Generative Adversarial Networks, Alec Radford etc., arXiv: 1511.06434, 2016

[14] Learning internal representations by error propagation. D E Rumellhart, G E Hinton，R J Williams. Proc. PDP, 1986: 318-362.

[15] Very Deep Convolutional Networks for Large-Scale Image Recognition, K. Simonyan etc. arXiv: 1409.1556

[16] Gradient-based learning applied to document recognition, Y. LeCun etc., Proc. IEEE, 1998, 86(11): 2278-2324

[17] Image semantic segmentation based on higher-order CRF model, L. Mao etc..Application Research of Computers, Nov. 2013, Vol. 30 No. 11

[18] Super pixel-based conditional random field for image classification, W. Zhang etc., Journal of Computer Applications, 2012, 32(5),1272-1275

[19] Rethinking the Inception Architecture for Computer Vision, C. Szegedy etc., Dec. 2015, http://arxiv.org/abs/1512.00567

[20] Going Deeper with Convolutions, C. Szegedy etc., Sep. 2014, http://arxiv.org/abs/1409.4842

[21] Batch Normalization: Accelerating Deep Network Training by Reducing Internal Covariate Shift, c. Szegedy etc., Mar. 2015, http://arxiv.org/abs/1502.03167